\documentclass[conference]{IEEEtran}
\IEEEoverridecommandlockouts
\usepackage{pifont}
\usepackage{booktabs}
\usepackage{cite}
\usepackage{amsmath,amssymb,amsfonts}
\usepackage{algorithmic}
\usepackage{graphicx}
\usepackage{tabularx}
\usepackage{textcomp}
\usepackage{xcolor}
\usepackage[table]{xcolor}  
\usepackage{colortbl}       
\usepackage{hyperref}
\hypersetup{hidelinks}
\usepackage[ruled,vlined]{algorithm2e}
\def\BibTeX{{\rm B\kern-.05em{\sc i\kern-.025em b}\kern-.08em
    T\kern-.1667em\lower.7ex\hbox{E}\kern-.125emX}}

\begin{document}

\title{OracleAnalyser: Analysing Implicit Semantics of Oracle Bone Scripts through MLLMs \\ with Post-training}

\author{
Zijia Song$^{1,*}$, Yelin Wang$^{2,*}$, Zhengyi Ma$^{1}$, and Zitong Yu$^{3}$\\
Tianheng Wang$^{4}$, Jiahuan Zhang$^{4}$, Taorui Wang$^{3}$, and Kaicheng Yu$^{4,\dagger}$\\
\small $^{1}$National University of Defense Technology \qquad $^{2}$ShanghaiTech University\\
\small $^{3}$Great Bay University \qquad $^{4}$Westlake University\\
\small songzijia@nudt.edu.cn, wangyl2023@shanghaitech.edu.cn, mazhengyi@nudt.edu.cn,\\
\small yuzitong@gbu.edu.cn, wangtianheng@him.cas.cn, zhangjiahuan78@westlake.edu.cn,\\
\small wangtaorui811@gmail.com, kyu@westlake.edu.cn\\
\small $^{*}$Equal contribution \quad $^{\dagger}$Corresponding author
}

\maketitle

\begin{abstract}
With the advancement of artificial intelligence, research on oracle bone scripts has entered a new era. However, existing methods and benchmarks remain largely confined to recognition tasks, overlooking the equally crucial aspect of oracle bone analysis. To address this gap, we propose OracleAnalyser, a reasoning framework for oracle bone analysis based on post-training techniques. Specifically, we fine-tune Qwen2.5-VL-3B-Instruct through multiple post-training stages and introduce a new preference optimization algorithm, Stable Focal Preference Optimization (SFPO), tailored to the characteristics of oracle bone datasets. In addition, we release both an oracle bone reasoning dataset and an oracle bone preference dataset, and further construct a new benchmark to evaluate models’ analytical capabilities for oracle bone scripts. Extensive experiments validate the superior analytical performance of OracleAnalyser, which achieves remarkable results with only 3B parameters, surpassing models with substantially larger scales.
\end{abstract}

\begin{IEEEkeywords}
oracle bone script analysis, multimodal large language model, post-training
\end{IEEEkeywords}

\section{Introduction}
\label{sec:intro}

Oracle Bone Script (OBS) is the earliest known mature writing system in China, originating in the Shang dynasty. These inscriptions were primarily engraved on turtle plastrons and animal bones for divinatory purposes. As a pictographic writing system, OBS recorded various aspects of ancient society and daily life, and provides invaluable evidence for understanding the language, culture, and history of early Chinese civilization. To date, more than $4500$ oracle bone characters have been identified, yet only about $1600$ have been deciphered and linked to modern Chinese characters, leaving the majority still undeciphered \cite{wang2024dataset}. The analysis and interpretation of these unknown characters are crucial for cultural heritage preservation and are of great significance to archaeology, historical studies, and philology. Consequently, oracle bone analysis has become an important and active research topic in the academic community \cite{jiang2023oraclepoints,OBS}.


Recent studies have begun applying modern AI techniques to oracle bone script decipherment. For example, OBSD \cite{OBSD} adopts a conditional diffusion model to capture font evolution patterns, while Puzzle Pieces Picker \cite{P3} employs Transformers to reconstruct strokes and decode characters. However, most existing methods primarily focus on recognition and overlook deeper analytical understanding. Similarly, current benchmarks \cite{OBI-Bench, HUST-OBC} are largely recognition-oriented and lack evaluation metrics for analytical capability. We argue that recognition mainly reflects memorization, whereas analytical competence is critical for effective oracle bone script decipherment.


\begin{figure}[tbp]
\centerline{\includegraphics[width=1\columnwidth]{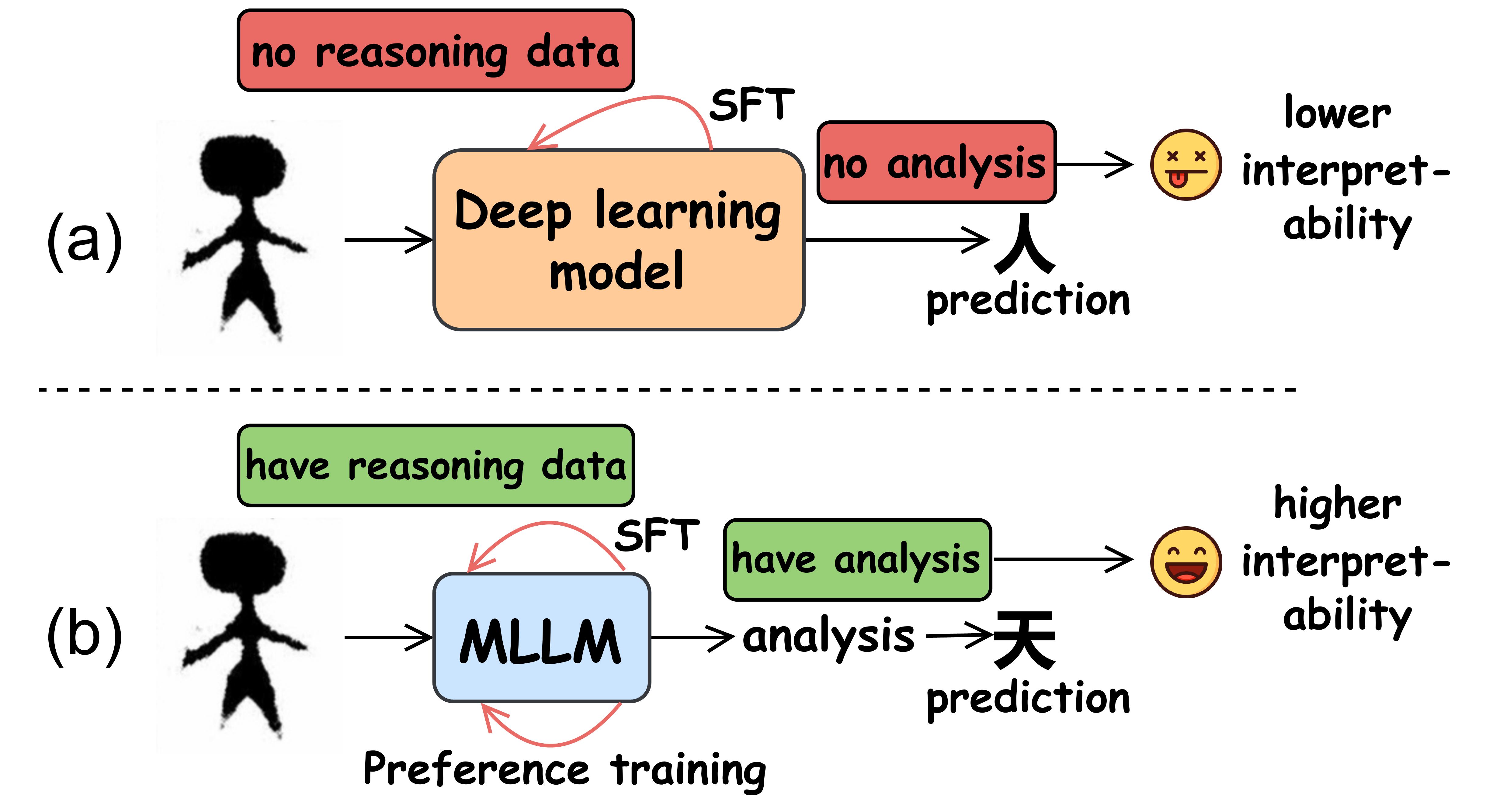}}
\caption{The differences between the previous approaches and ours. (a) Previous approaches can not perform analytical reasoning, resulting in limited interpretability. (b) our method can generate analysis before recognition, making it more conducive to the decipherment of oracle bone scripts.}
\label{intro}
\vspace{-0.5cm}
\end{figure}

Post-training techniques combined with reasoning have led to significant breakthroughs in general domains \cite{tie2025survey, deepseek-r1, Wang2025InternVL35, zhang2025acl}. 
These achievements inspire us to consider: if the analysis of oracle bone characters can be regarded as a specific form of reasoning, could such post-training approaches be leveraged to facilitate their interpretation and decipherment? Motivated by this idea, we propose OracleAnalyser, a reasoning framework for oracle bone analysis based on post-training techniques. Specifically, we construct a new oracle bone reasoning dataset and use it to fine-tune Qwen2.5-VL-3B-Instruct \cite{qwen2_5_vl} under supervised learning, obtaining OracleAnalyser-sft. We then employ OracleAnalyser-sft for inference to generate positive and negative sample pairs, thereby building a preference dataset tailored for oracle bone analysis. Furthermore, we propose a new algorithm for oracle bone analysis, Stable Focal Preference Optimization (SFPO), to further train OracleAnalyser-sft, resulting in the final OracleAnalyser. The differences between our method and previous deep learning approaches are illustrated in Fig.~\ref{intro}.
Meanwhile, we design a new benchmark to evaluate the intermediate analytical reasoning output of MLLMs rather than focusing only on recognition outcomes. 


\begin{figure*}[tbp]
\centerline{\includegraphics[width=0.95\textwidth]{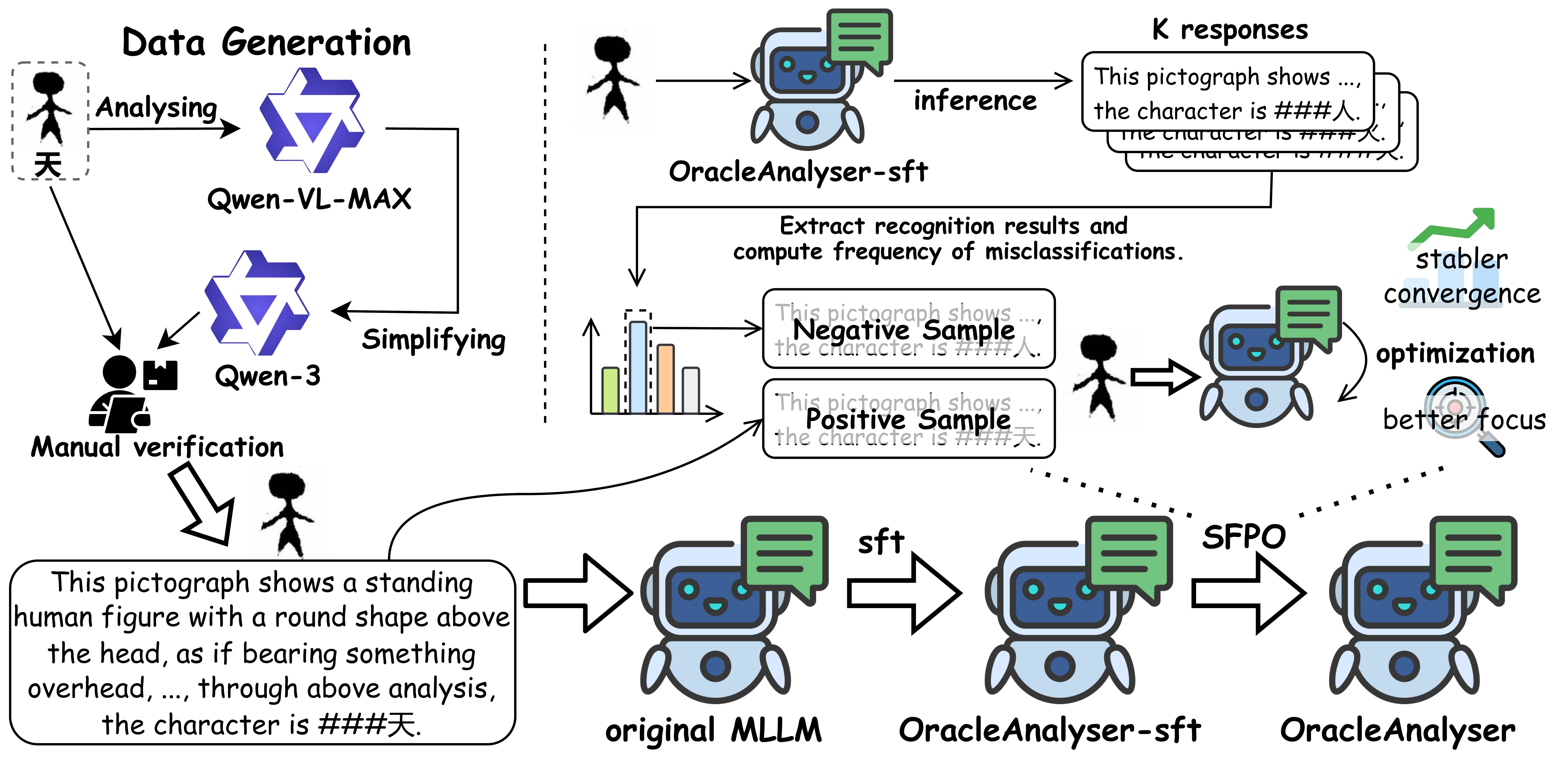}}
\vspace{-0.5cm}
\caption{The overall framework of OracleAnalyser. It employs reasoning combined with post-training techniques to analyse and recognize oracle bone scripts. The original MLLM is Qwen2.5-VL-3B-Instruct. The post-training process consists of supervised fine-tuning (SFT) and the proposed SFPO which can achieve stabler convergence and provides better focus on valuable samples.}
\label{framework}
\vspace{-0.3cm}
\end{figure*}

Our contributions can be summarized as follows:
\begin{itemize}
    \item We release a newly constructed \textbf{oracle bone reasoning dataset} and the \textbf{first oracle bone preference dataset}. In addition, we design \textbf{a new benchmark to systematically evaluate oracle bone analysis}.

    \item We propose a novel oracle bone analysis framework, named \textbf{OracleAnalyser}, and introduce a new preference optimization objective, \textbf{Stable Focal Preference Optimization (SFPO)}, tailored to the characteristics of the oracle bone dataset. To the best of our knowledge, this is the first work that integrates explicit reasoning with multiple post-training strategies for oracle bone analysis.

    \item We conduct extensive experiments, and the results demonstrate that \textbf{OracleAnalyser significantly outperforms existing MLLMs} on oracle bone analysis.
\end{itemize}



\section{Related works}

\noindent\textbf{Intelligent recognition and decipherment of oracle-bone scripts.} 
For benchmarks, EVOBC compiles multi-era glyphs for evolution-aware decipherment \cite{EVOBC}, and HUST-OBC provides a large-scale corpus and benchmark \cite{HUST-OBC}. OBI-Bench evaluates LMMs end to end on oracle-bone tasks, exposing gaps in fine-grained perception and domain knowledge \cite{OBI-Bench}. For methods, $P^3$ reconstructs radicals to turn undeciphered characters into analysable sequences \cite{P3}, and OracleSage employs the MLLM and fuses hierarchical visual structure with knowledge-graph reasoning \cite{OracleSage}. CFIRN casts decipherment as cross-font or cross-era retrieval with a Siamese, multi-scale model \cite{CFIRN}, and OBSD utilizes a conditional diffusion-based strategy for generating vital clues for decipherment \cite{OBSD}. However, current benchmarks and methods mainly focus on recognition and often neglect oracle bone analysis.

\noindent\textbf{The reasoning progress of MLLMs.}
Recently, reinforcement fine-tuning (RFT) has significantly advanced the reasoning capabilities of MLLMs \cite{wang2025simplear,li2024look,liu2025noisyrollout,chen2025compile,wang2025skywork, zhang2025adamhf}. Li et al. propose CLS-RL and No-Thinking-RL, and argue that explicit reasoning is not always beneficial \cite{Li2025ThinkOrNotThink}. At scale,  Kimi-VL integrates Mixture-of-Experts (MoE), long-context modeling, and long-CoT supervision to enhance multimodal reasoning \cite{KimiTeam2025KimiVL,KimiTeam2025KimiK15}. InternVL-3.5 adopts a cascaded reinforcement learning framework, reporting substantial gains across diverse multimodal reasoning benchmarks \cite{Wang2025InternVL35}. Along the R1-style research line, VLM-R1 adapts the R1 paradigm to vision-language models, achieving better generalization than standard SFT \cite{Shen2025VLMR1}. Similarly, Perception-R1 demonstrates perception-policy improvements on COCO and related benchmarks \cite{Yu2025PerceptionR1}, VisRL directly optimizes the intention-attention-reasoning loop via reinforcement learning \cite{Chen2025VisRL}, and OpenVLThinker employs iterative SFT-RL to transfer text-based reasoning paradigms into multimodal contexts \cite{Deng2025OpenVLThinker}. The current progress of MLLMs also provides new solutions for oracle bone analysis.


\section{Method}
\label{sec:method}

\subsection{Overview of Framework}

Robust oracle bone analysing capabilities are crucial for deciphering these enigmatic symbols \cite{v-oracle}. Motivated by the reasoning progress in general domains, we propose OracleAnalyser, a reasoning framework for oracle bone analysis based on post-training methods. The overall framework is illustrated in Fig.~\ref{framework}. Specifically, we first prompt Qwen-VL-Max to construct an oracle bone reasoning dataset, which is then to supervise Qwen2.5-VL-3B-Instruct via autoregressive learning, yielding OracleAnalyser-SFT. The trained model is subsequently used to generate multiple inference traces for each oracle bone image, from which positive and negative samples are constructed to form the oracle bone preference dataset. To overcome the limitations of directly applying DPO to oracle bone analysis, we further propose Stable Focal Preference Optimization (SFPO) and apply it to preference training, resulting in the final OracleAnalyser model.


\begin{figure}[tbp]
\centerline{\includegraphics[width=0.95\columnwidth]{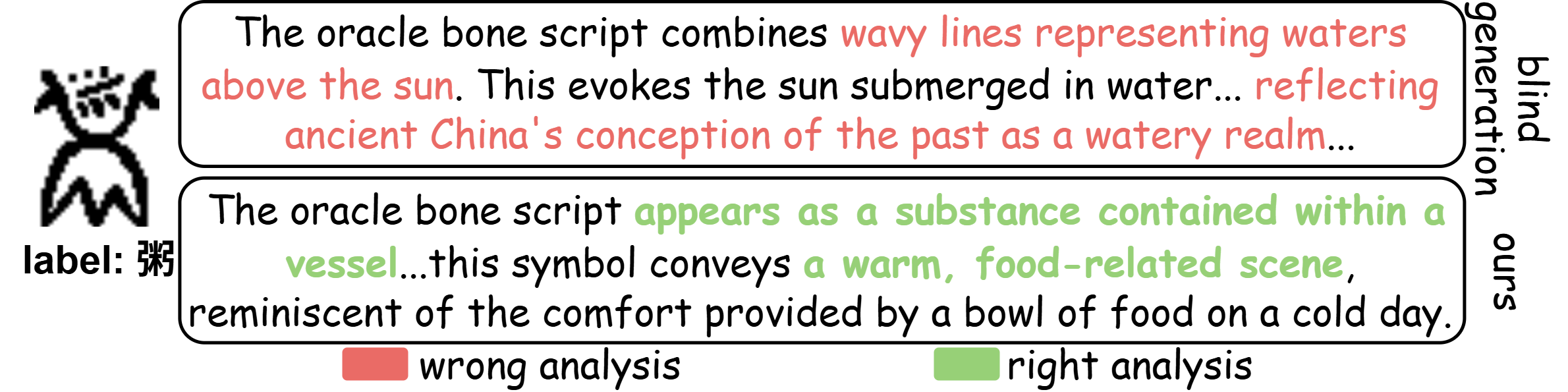}}
\vspace{-0.3cm}
\caption{Visualization comparison between blind generation (without modern character hints) and our generation paradigm.}
\label{whether_blind_generation}
\vspace{-0.3cm}
\end{figure}

\subsection{Construction of Oracle Bone Reasoning Dataset}
\label{sec:3.2}
\begin{figure}[tbp]
\centerline{\includegraphics[width=0.95\columnwidth]{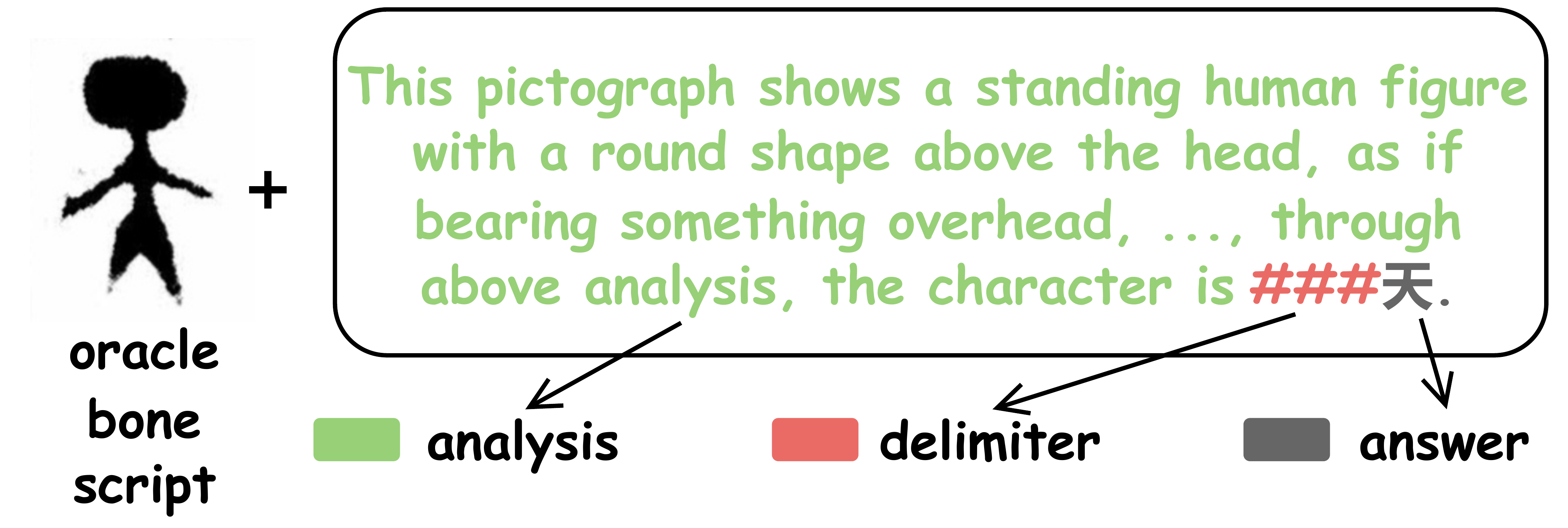}}
\vspace{-0.3cm}
\caption{The format of a sample in oracle bone reasoning dataset.}
\label{data_format}
\vspace{-0.5cm}
\end{figure}

We construct our dataset based on the OBC subset of the Evolution Dataset of Oracle Bone Characters (EVOBC), where each sample pairs an oracle bone image with its corresponding modern Chinese character. Since EVOBC contains obsolete characters, we filter the data using a list of $3500$ commonly used Chinese characters, resulting in $2807$ valid characters. To mitigate class imbalance caused by multiple oracle bone images per character, we cap the number of images per character at $10$. 
To generate analytical descriptions, we input each oracle bone image into Qwen-VL-MAX \cite{qwen2_5_vl} using the prompt:
\texttt{This is the oracle bone of {hz}. Describe the script from a pictographic perspective, but do not mention {hz} explicitly in your description},
where \texttt{{hz}} denotes the corresponding modern Chinese character. Conditioning on the modern character enables more accurate analysis of the oracle bone script, as shown in Fig. \ref{whether_blind_generation}. The generated descriptions are then refined by Qwen-3 \cite{qwen3} to remove redundancy using the prompt:
\texttt{For the analysis of an oracle bone script {analysis}, please help me make it more concise.}
Finally, all samples undergo manual verification to prevent data leakage and eliminate low-quality samples, yielding approximately $6000$ samples. We retain $5500$ samples in total, with $5000$ used for training, $500$ for in-domain testing. The remaining $500$ samples are reserved as an out-of-domain test set, whose modern counterparts are entirely absent from the training set, enabling evaluation of the model's generalization capability. The manual verification process involves two professional researchers and several paleography students. Each data entry consists of an oracle bone image, its analytical description, and the corresponding modern Chinese character, as illustrated in Fig.~\ref{data_format}.

\subsection{Supervised Fine-tuning Process}
\label{sec:3.3}
We adopt Supervised Fine-tuning (SFT) as the first stage of post-training. In this stage, the model could gradually captures the intrinsic visual patterns of oracle bone scripts, and develop fundamental reasoning abilities for analysing oracle bone characters. Moreover, this stage provides a stable initialization for subsequent preference training, serving as a critical bridge between pre-trained knowledge and specialized reasoning abilities required for oracle bone analysis. 

Specifically, we employ Qwen2.5-VL-3B-Instruct as backbone, and the image of oracle bone character, together with the prompt: \texttt{This is an oracle bone character. Please try to analyse it and identify it, and use \#\#\# to separate the analysis from the recognition result}, are fed into the MLLM to generate structured analyses of the symbols. During training, the model’s output is compared against the ground-truth, and the autoregressive loss is introduced for optimization. The formula is shown as follows:
\begin{equation}
\mathcal{L}_{AR} = - \sum_{t=1}^{T} \log P\!\left(y_t \mid y_{<t}, \,I, \,P; \theta \right),
\end{equation}
where $I$ denotes the visual embedding from the vision encoder, $P$ is the prompt token, $y_t$ is the $t$-th output token, $T$ is the sequence length, and $\theta$ represents model parameters. After SFT, we obtain the model OracleAnalyser-SFT. This stage equips the model with fundamental recognition and interpretative abilities, thereby facilitating effective and stable preference optimization in subsequent stages.



\vspace{-0.2cm}
\subsection{Stable Focal Preference Optimization}
\label{sec:3.4}
Preference learning is a widely used post-training technique that improves generation quality by exploiting relative preferences rather than explicit labels. In oracle bone analysis, it can further refine the analytical capability of OracleAnalyser-SFT.

The effectiveness of preference learning critically depends on the quality of preference pairs. To construct high-quality data, we input each oracle bone image and its prompt into OracleAnalyser-SFT and perform five inference traces. Recognition results are extracted using delimiters, and incorrect predictions are grouped by frequency. Formally, we obtain ${\mathbf{s}_i, |\mathbf{s}_i|}_i$, where $\mathbf{s}_i$ denotes the set of reasoning traces producing the same incorrect recognition and $|\mathbf{s}_i|$ its frequency. From the set with the highest $|\mathbf{s}_i|$, we randomly select one trace as the negative sample, while the ground-truth analysis serves as the positive sample, forming a preference pair.

Applying this procedure to all oracle bone images yields over $4700$ preference pairs. After manual filtering to remove ambiguous or low-contrast cases, we retain more than $3400$ high-quality pairs as the oracle bone preference dataset. By selecting negatives that reflect the model’s own systematic errors and discarding trivial cases, this dataset enables effective optimization of analytical reasoning.


Direct Preference Optimization (DPO) is a widely adopted algorithm in preference learning. Let $\mathcal{D} = {(x^{(i)}, y_{w}^{(i)}, y_{l}^{(i)})}_{i=1}^{N}$ represent the oracle bone preference dataset, where $x^{(i)} \in \mathcal{X}$ consists of a prompt and the corresponding oracle bone image. Each $(y_{w}^{(i)}, y_{l}^{(i)})$ denotes a pair of candidate answers, with the preference of $y_{w}^{(i)} \succ y_{l}^{(i)}$. Following the Bradley–Terry model, the preference distribution $p^{*}$ can be formulated as:
\begin{equation}
p^{*}(y_{w}^{(i)} \succ y_{l}^{(i)} \mid x^{(i)}) = 
\sigma\big(r^{*}(x^{(i)}, y_{w}^{(i)}) - r^{*}(x^{(i)}, y_{l}^{(i)})\big),
\end{equation}
where $\sigma$ is the sigmoid function and $r^{*}(x, y)$ is the latent reward model used to generate the ground-truth preference. 
To model the preference, DPO uses a reparametrization trick to express it in terms of the optimal policy $\pi^*$:
\begin{equation}
    r^{*}(x,y) = \beta \log \frac{\pi^{*}(y \mid x)}{\pi_{\text{ref}}(y \mid x)} + \beta \log Z(x),
    \label{r*}
\end{equation}
where $Z(x)$ is the partition function only based on $x$.  
Through equation~\ref{r*}, the maximum likelihood estimation objective of DPO is as follows:
\begin{small}
\begin{equation}
\begin{aligned}
& \mathcal{L}_{\text{DPO}}(\pi_\theta; \pi_{\text{ref}}) 
= - \mathbb{E}_{(x,y_w,y_l) \sim \mathcal{D}} \left[ \log p(y_w \succ y_l \mid x) \right] =\\
& - \mathbb{E}_{(x,y_w,y_l) \sim \mathcal{D}} \Bigg[ 
\log \sigma \Bigg( 
\beta \log \frac{\pi_\theta(y_w \mid x)}{\pi_{\text{ref}}(y_w \mid x)} - \beta \log \frac{\pi_\theta(y_l \mid x)}{\pi_{\text{ref}}(y_l \mid x)} 
\Bigg) \Bigg],
\end{aligned}
\end{equation}
\end{small}
where minimizing $\mathcal{L}_{\text{DPO}}$ encourages $\pi_{\theta}$ to align with $\pi^{*}$, thereby increasing the quality of producing responses.

However, directly applying DPO in oracle bone analysis is suboptimal. The preference dataset contains noisy positive samples and partially correct negatives, which can cause the policy model to overfit imperfect preferences and drift excessively from the reference model, violating the assumptions of DPO and leading to reward hacking. In addition, DPO tends to emphasize learning from response pairs that the model struggles to rank. In the oracle bone preference dataset, many such hard pairs contain partially correct or reasonable content on both sides. Thus, DPO training with enforcing a strict preference may exaggerate minor differences, leading the model to over-interpretation or hallucinated distinctions.


To address these issues, we propose Stable Focal Preference Optimization (SFPO), a variant of DPO tailored for oracle bone analysis. SFPO enforces controlled divergence between the policy and reference models to stabilize training, while down-weighting hard-to-rank preference pairs and prioritizing pairs with reliable implicit reward signals. Under these principles, the loss function of SFPO is formulated as follows:
\begin{equation}
  \mathcal{L}_{\text{SFPO}}(\pi_\theta; \pi_{\text{ref}}) = \mathcal{L}_{\text{F}}(\pi_\theta; \pi_{\text{ref}}) + \lambda \mathcal{L}_{\text{S}}(\pi_\theta; \pi_{\text{ref}}), 
\end{equation}
where $\pi_\theta$ denotes the model continuously optimized during post-training, while $\pi_{\text{ref}}$ refers to OracleAnalyser-sft. The objective of $\mathcal{L}_{\text{F}}(\cdot; \cdot)$ aims to guide the model to reduce excessive attention to hard-to-rank pairs. To achieve this, FocalPO \cite{focalpo} is introduced as $\mathcal{L}_{\text{F}}(\cdot; \cdot)$, which is defined as follows:
\begin{equation}
\begin{aligned}
\mathcal{L}_{\text{F}}(&\pi_\theta;\pi_{\text{ref}}) = -\mathbb{E}_{(x,y_w,y_l)\sim\mathcal{D}} 
\big[\\ &  p(y_w \succ y_l \mid x)^{\gamma}\log p(y_w \succ y_l \mid x) \big] + \text{c},
\end{aligned}
\end{equation}
where $\gamma$ is a tunable focusing hyperparameter set to $0.05$, and $\text{c}$ denotes a constant term. When the implicit reward model correctly classifies a preference pair, the modulating factor is close to $1$ and leaves the loss unchanged. Conversely, as $p(y_w \succ y_l \mid x) \to 0$, the factor approaches $0$, thereby reducing its contribution to the objective. $\mathcal{L}_{\text{S}}(\cdot; \cdot)$, on the other hand, stabilizes training by constraining the divergence between the $\pi_\theta$ and $\pi_{ref}$. The formula is shown as follows:
\begin{equation}
\mathcal{L}_{\text{S}}(\pi_\theta;\pi_{\text{ref}}) = \mathbb{E}_{x\sim\mathcal{D}} \left[ D_{\mathrm{KL}}\left( \pi_\theta(\cdot \mid x) \,\|\, \pi_{\text{ref}}(\cdot \mid x) \right) \right],
\end{equation}
here, we explicitly introduce the reverse KL divergence as stabilizing term to prevent the policy model from deviating excessively from the reference model, thus ensuring stable training. Guided by SFPO, the model can focus on more valuable samples during the preference training stage and achieve stabler optimization, ultimately yielding OracleAnalyser. Moreover, SFPO does not consistently assign low weights to hard pairs, and it learns to distinguish them and gradually increases their attention weights during training, thereby learning informative signals at the opportune time.





\setlength{\tabcolsep}{4pt}
\begin{table}[tbp]
\centering
\caption{Performance comparison of different models on the in-domain test set to evaluate the oracle bone analysis and recognition capabilities. Acc denotes recognition accuracy, and the best results are highlighted in \textbf{bold}.}
\scalebox{1}{
\begin{tabular}{lcccccc}
\toprule
Method & BLEU-1 & BLEU-4 & ROUGE-1 & ROUGE-L & Acc \\
\midrule
Qwen-VL-MAX        & 29.11 & 4.43 & 32.41 & 20.14 & 2.4 \\
GLM-4.5V           & 28.11 & 3.65 & 31.68 & 20.02 & 3.2 \\
SenseChat-Vision   & 27.91 & 3.70 & 31.87 & 19.68 & 3.8 \\
Intern-S1          & 28.32 & 3.89 & 31.58 & 20.82 & 14.0 \\  \hline
InternVL-3.5-241B  & 28.21 & 3.84 & 31.55 & 20.78 & 12.8 \\
InternVL-3-78B     & 27.79 & 3.54 & 31.50 & 19.91 & 7.2 \\
Qwen3-VL-30B       & 28.41 & 3.44 & 32.05 & 18.94 & 1.2 \\
Qwen2.5-VL-72B     & 29.61 & 4.32 & 32.30 & 20.02 & 1.6 \\
Qwen2.5-VL-7B      & 20.77 & 3.07 & 26.92 & 16.52 & 0.6 \\ \hline
BBDM               &  -  &  -   &   -    & -  & 16.9 \\
OBSD               &  -  &  -   &   -    & -  & 38.7 \\
\hline
Qwen2.5-VL-3B      & 18.22 & 2.23 & 25.78 & 15.60 & 0.4 \\
\rowcolor{yellow!20}
OracleAnalyser-sft & 86.43 & 69.86 & 64.68 & 62.05 & 36.2 \\
\rowcolor{yellow!20}
OracleAnalyser     & \textbf{87.92} & \textbf{72.77} & \textbf{67.86} & \textbf{65.42} & \textbf{39.6} \\
\bottomrule
\end{tabular}}
\label{tab:contrast-in-domain}
\vspace{-0.2cm}
\end{table}

\begin{table}[tbp]
\centering
\caption{Performance comparison of different models on the out-of-domain test set to evaluate the analytical capability when facing unseen oracle bone characters. The best results are highlighted in \textbf{bold}.}
\setlength{\tabcolsep}{4pt} 
\scalebox{1}{
\begin{tabular}{lccccc}
\toprule
Method & BLEU-1 & BLEU-4 & ROUGE-1 & ROUGE-L & SBS \\
\midrule
Qwen-VL-MAX        & 29.18 & 4.45 & 32.37 & 20.16 & 69.98 \\
GLM-4.5V           & 28.09 & 3.67 & 31.65 & 19.93 & 68.92 \\
SenseChat-Vision   & 27.88 & 3.69 & 31.89 & 19.65 & 69.02 \\
Intern-S1          & 28.37 & 3.86 & 31.58 & 20.83 & 68.88 \\ \hline
InternVL-3.5-241B  & 28.27 & 3.84 & 31.58 & 20.81 & 68.81 \\
InternVL-3-78B     & 27.83 & 3.39 & 31.06 & 19.49 & 67.92 \\
Qwen3-VL-30B       & 28.47 & 3.32 & 31.48 & 18.57 & 60.34 \\
Qwen2.5-VL-72B     & 28.63 & 4.35 & 32.42 & 20.04 & 68.75 \\
Qwen2.5-VL-7B      & 20.36 & 3.04 & 26.85 & 16.33 & 58.97 \\ \hline
Qwen2.5-VL-3B      & 18.81 & 2.08 & 25.08 & 15.22 & 49.99 \\
\rowcolor{yellow!20}
OracleAnalyser-sft & 76.77 & 49.54 & 38.81 & 33.82 & 69.22 \\
\rowcolor{yellow!20}
OracleAnalyser     & \textbf{77.61} & \textbf{50.13} & \textbf{39.85} & \textbf{35.01} & \textbf{70.35} \\
\bottomrule
\end{tabular}}
\label{tab:contrast-out-of-domain}
\vspace{-0.3cm}
\end{table}

\section{Experiments}

\subsection{Proposed Benchmark}

Existing benchmarks focus mainly on recognition and fail to evaluate the analytical capability required for oracle bone script decipherment. To address this gap, we propose a new benchmark for assessing the reasoning ability of MLLMs in oracle bone analysis, comprising both in-domain and out-of-domain evaluations.

The in-domain evaluation assesses fundamental reasoning and recognition performance on an in-domain test set of 500 samples. To measure analytical quality, we report BLEU-1, BLEU-4, ROUGE-1, and ROUGE-L, where BLEU-1 and ROUGE-1 reflect key-term accuracy, while BLEU-4 and ROUGE-L capture sentence-level coherence. Recognition accuracy is additionally reported to evaluate final identification performance.
The out-of-domain evaluation examines generalization to unseen oracle bone scripts using an out-of-domain test set of approximately 500 samples. In order to more accurately reflect the model’s analytical capability on unseen characters, recognition accuracy is replaced by Sentence-BERT Score (SBS), defined as
\begin{equation}
\text{SBS} = \frac{1}{N} \sum_{i=1}^N \text{cosine}(p_i, l_i),
\end{equation}
where $N$ denotes the number of samples in the dataset, and cosine represents the cosine similarity. $p_i$ and $l_i$ are the Sentence-BERT embeddings of the model output and ground-truth analysis. A higher SBS indicates the model realizes stronger analytical capability on unseen characters.



\begin{figure}
    \centering
    \includegraphics[width=1\linewidth]{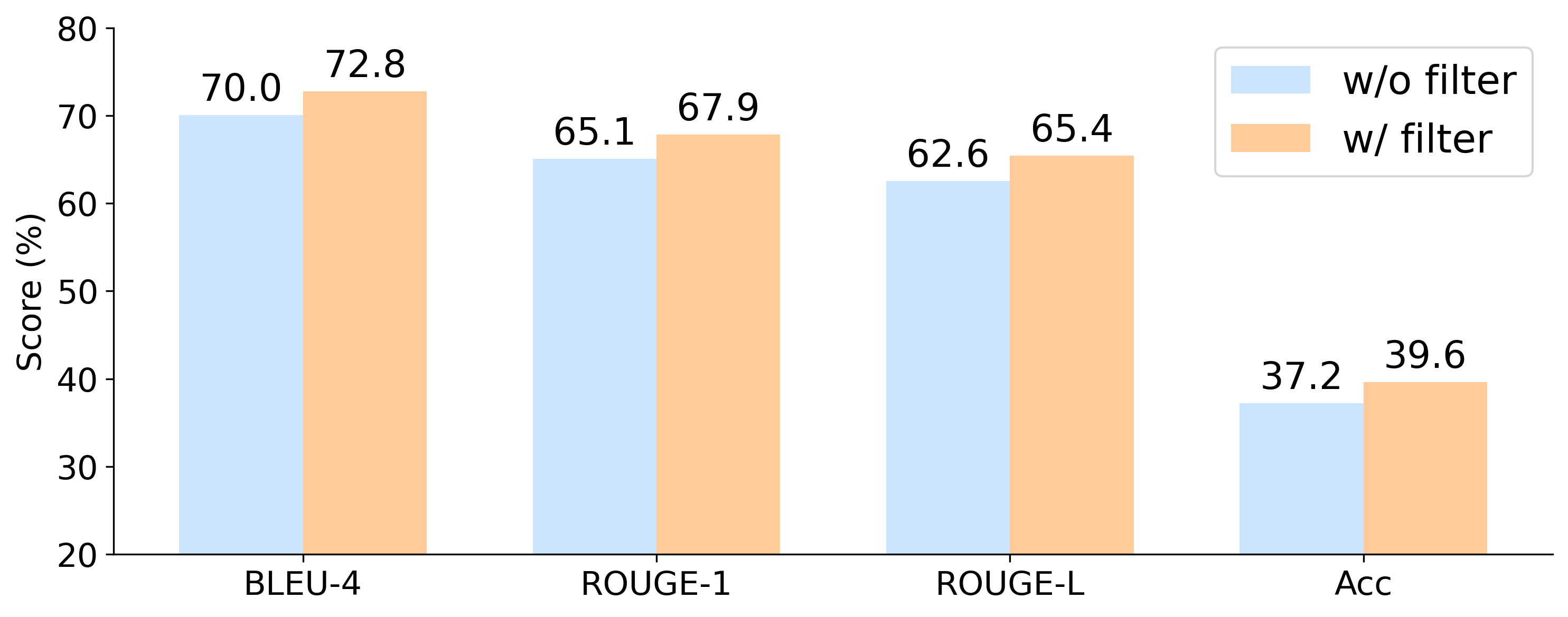}
    \caption{Ablation study on whether to filter the preference data.}
    \label{fig:data_filter}
    \vspace{-0.4cm}
\end{figure}

\begin{table}[tbp]
\centering
\caption{Ablation study on different training strategies.}
\setlength{\tabcolsep}{5pt} 
\scalebox{1}{
\begin{tabular}{lcccc}
\toprule
Training & BLEU-4 & ROUGE-1 & ROUGE-L & Acc \\
\midrule
SFT          & 69.86 & 64.68 & 62.05 & 36.2 \\
SFT+DPO      & 70.84 & 65.77 & 63.19 & 39.4 \\
SFT+$\mathcal{L}_{\text{F}}$     & 72.46 & 67.65 & 65.09 & \textbf{39.8} \\
SFT+$\mathcal{L}_{\text{F}}$+$\mathcal{L}_{\text{S}}$ & \textbf{72.77} & \textbf{67.86} & \textbf{65.42} & 39.6 \\
\bottomrule
\end{tabular}}
\label{tab:training_strategies}
\vspace{-0.3cm}
\end{table}

\subsection{Implementation Details}

In our experimental setup, OracleAnalyser-sft is fine-tuned on top of Qwen2.5-VL-3B with LoRA rank set to $8$. The batch size is $32$, and the model is trained for $100$ epochs with a learning rate of 5e-5.
In the SFPO stage, the model continues training based on OracleAnalyser-sft, with a batch size of $8$ and $20$ training epochs. The learning rate is set to 5e-6, and the balancing coefficient $\lambda$ is $0.5$. In $\mathcal{L}_{\text{F}}$, the parameters $\beta$ and $\gamma$ are set to $0.1$ and $0.05$, respectively.



\subsection{In-domain and Out-of-domain Evaluation}

We compare OracleAnalyser with other competitive models on both in-domain and out-of-domain test sets. Except for Qwen2.5-VL-3B (our baseline), all compared MLLMs have substantially larger parameter scales. BBDM and OBSD are trained specialized models for oracle bone recognition.
In-domain results are reported in TABLE~\ref{tab:contrast-in-domain}. OracleAnalyser consistently outperforms all competing models across all metrics. It achieves improvements of 35\%–70\% on analysis metrics and a higher recognition accuracy over specialized models. 
Out-of-domain results, shown in TABLE~\ref{tab:contrast-out-of-domain}, demonstrate strong generalization on unseen characters. OracleAnalyser surpasses other models by 10\%–50\% across metrics and achieves the highest SBS score. This indicates that its generated analysis are more semantically aligned with ground truth and more effective for oracle bone decipherment. For practical applications, OracleAnalyser can provide reliable analysis when facing unknown scripts, thereby helping paleographers narrow the search space. Overall, these results confirm that OracleAnalyser learns robust and generalizable reasoning capabilities with only 3B parameters.



\subsection{Ablation Study}

\noindent\textbf{Ablation study on whether to filter the preference data.}
We conduct an ablation study to assess the effect of preference data filtering, where ambiguous and hard-to-distinguish pairs are removed. By filtering, the scale of dataset is reduced from over 4700 pairs to approximately 3400. OracleAnalyser-SFT is trained with SFPO using both datasets, and the results are shown in Fig.~\ref{fig:data_filter}.
Model trained on the filtered dataset consistently outperforms the model trained on the unfiltered data across various metrics. These results indicate that, for oracle bone analysis, preference data quality is more important than quantity, and removing ambiguous pairs leads to more effective preference optimization.


\begin{figure}
    \centering
    \includegraphics[width=1\linewidth]{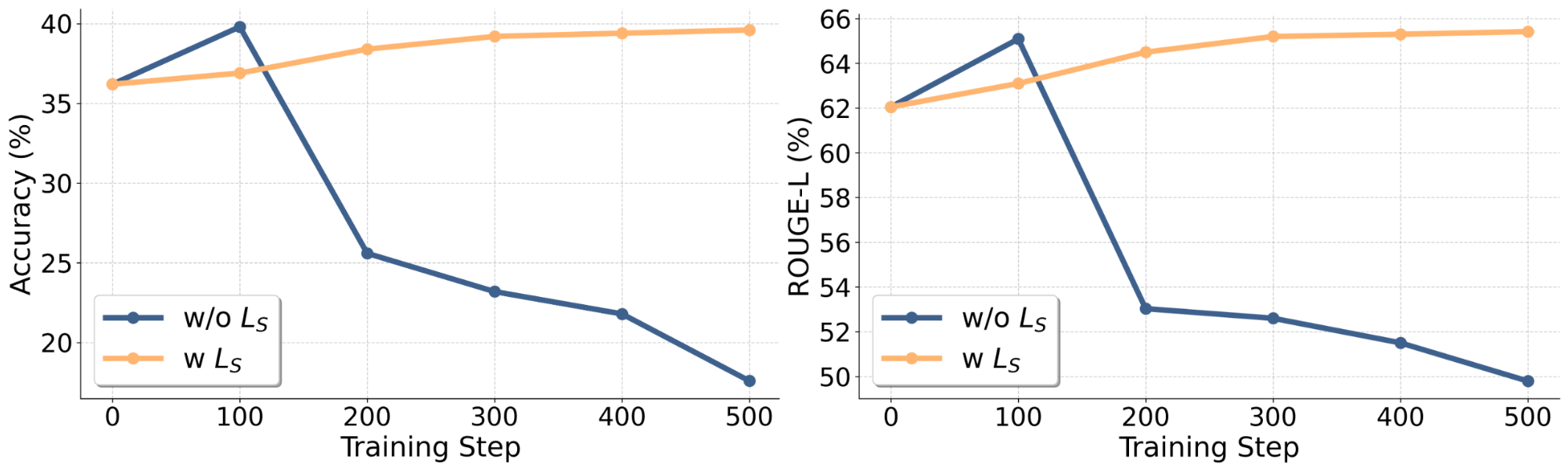}
    \caption{Ablation study on whether to employ $\mathcal{L}_{\text{S}}$}
    \label{fig:ablation_l_s}
    \vspace{-0.3cm}
\end{figure}

\begin{figure}[tbp]
    \centering
    \includegraphics[width=1\linewidth]{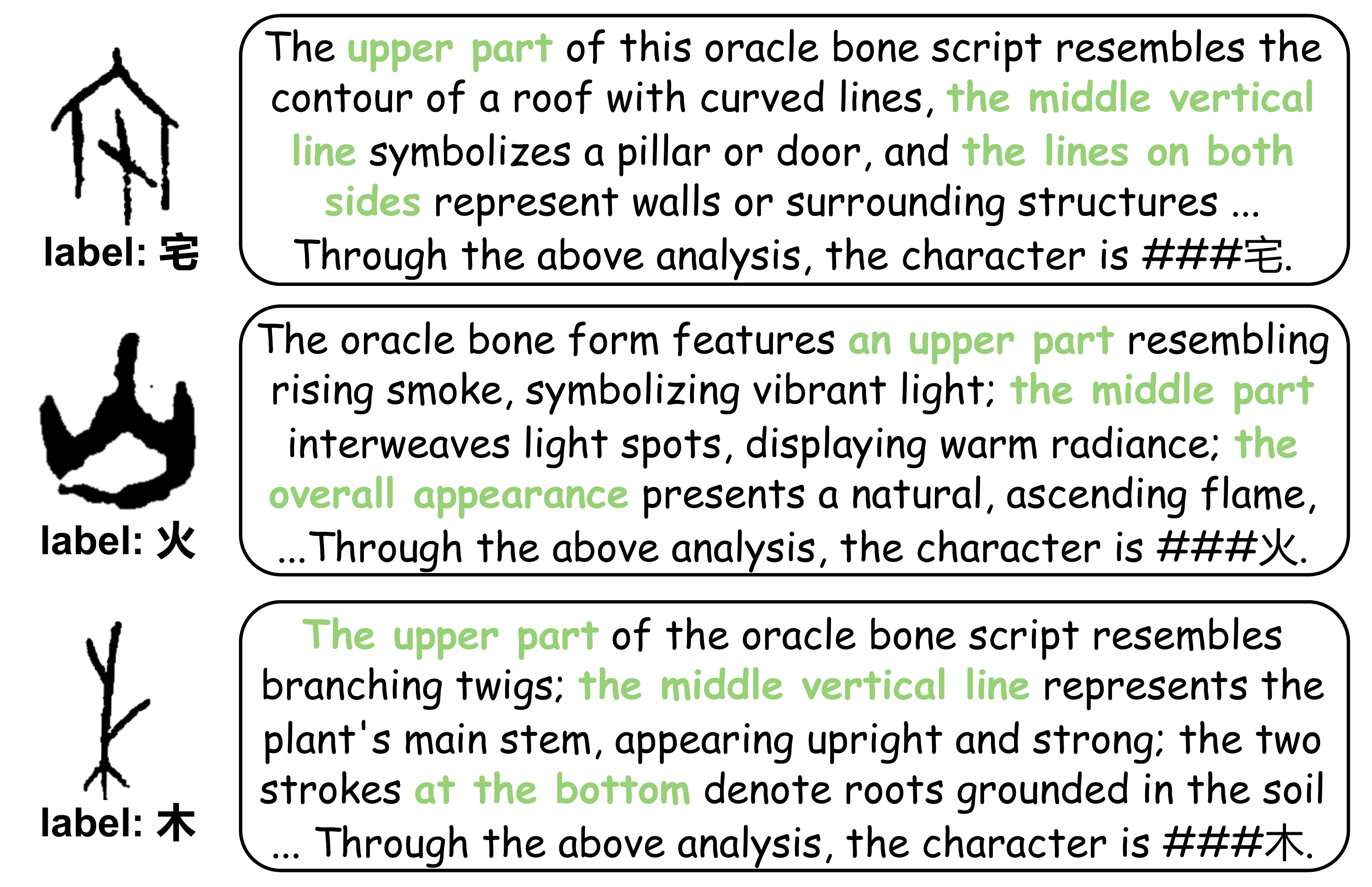}
    \caption{Visualization of OracleAnalyser outputs.}
    \label{fig:visualize}
    \vspace{-0.5cm}
\end{figure}

\noindent\textbf{Ablation study on different training strategies.}
We perform an ablation study on the in-domain test set to examine the impact of different training strategies, with results reported in TABLE~\ref{tab:training_strategies}. Applying DPO on top of SFT yields limited improvement, indicating that indistinguishable preference pairs negatively affect training. In contrast, $\mathcal{L}_{\text{F}}$ provides greater robustness by down-weighting the hard-to-rank pairs. Joint optimization with both $\mathcal{L}_{\text{F}}$ and $\mathcal{L}_{\text{S}}$ achieves the best averaged performance, validating the effectiveness of SFPO framework.

\noindent\textbf{Ablation study on whether to employ $\mathcal{L}_{\text{S}}$.}
We conduct an ablation study on the in-domain test set to evaluate the effect of $\mathcal{L}_{\text{S}}$ with recognition accuracy and ROUGE-L as metrics. As shown in Fig.~\ref{fig:ablation_l_s}, models trained without $\mathcal{L}_{\text{S}}$ show faster early gains but quickly degrade and eventually collapse. In contrast, incorporating $\mathcal{L}_{\text{S}}$ yields stable training and superior performance in later stages. These results demonstrate the effect of introducing a stabilizing term $\mathcal{L}_{\text{S}}$, mitigating the impact of noisy preferences and stabilizing optimization.


\subsection{Visual Analysis}

To better understand OracleAnalyser’s reasoning behavior, we visualize its outputs in Fig.~\ref{fig:visualize}. The results show that OracleAnalyser can accurately perceive and decompose oracle bone glyphs, and associate their components with real-world entities. This indicates that the model has learned to infer hidden semantics from pictographic structures. Evidently, through explicit reasoning and multi-stage post-training, OracleAnalyser demonstrates fundamental pictographic analysis capabilities which are valuable for oracle bone decipherment.



\section{Conclusion}
In this paper, we first identify the limitations of existing approaches and benchmarks in the oracle bone field, which focus mainly on recognition accuracy while neglecting equally crucial analytical reasoning. To address this, we propose OracleAnalyser, a post-training-based reasoning framework for oracle bone analysis. We also release an oracle bone reasoning dataset, an oracle bone preference dataset, and a new benchmark to evaluate the MLLM's analytical capabilities on oracle bone scripts. Extensive experiments demonstrate that OracleAnalyser achieves significantly superior analytical performance with only 3B parameters. 


\bibliographystyle{IEEEbib}
\bibliography{icme2026references}

\begin{thebibliography}{10}

\bibitem{wang2024dataset}
Mei Wang and Weihong Deng,
\newblock ``A dataset of oracle characters for benchmarking machine learning algorithms,''
\newblock {\em Scientific Data}, vol. 11, no. 1, pp. 87, 2024.

\bibitem{jiang2023oraclepoints}
Runhua Jiang, Yongge Liu, et~al.,
\newblock ``Oraclepoints: A hybrid neural representation for oracle character,''
\newblock in {\em Proceedings of the 31st ACM international conference on multimedia}, 2023, pp. 7901--7911.

\bibitem{OBS}
Kaixin Peng, Mengyang Zhao, et~al.,
\newblock ``Interpretable oracle bone script decipherment through radical and pictographic analysis with lvlms,''
\newblock {\em arXiv preprint arXiv:2508.10113}, 2025.

\bibitem{OBSD}
Haisu Guan, Huanxin Yang, et~al.,
\newblock ``Deciphering oracle bone language with diffusion model,''
\newblock in {\em Proceedings of the 62th Annual Meeting of the Association for Computational Linguistics}, 2024.

\bibitem{P3}
Pengjie Wang, Kaile Zhang, et~al.,
\newblock ``Puzzle pieces picker: Deciphering ancient chinese characters with radical reconstruction,''
\newblock in {\em International Conference on Document Analysis and Recognition}. Springer, 2024.

\bibitem{OBI-Bench}
Zijian Chen, Tingzhu Chen, et~al.,
\newblock ``Obi-bench: Can lmms aid in study of ancient script on oracle bones?,''
\newblock {\em arXiv preprint arXiv:2412.01175}, 2024.

\bibitem{HUST-OBC}
Pengjie Wang, Kaile Zhang, et~al.,
\newblock ``An open dataset for oracle bone script recognition and decipherment,''
\newblock {\em arXiv preprint arXiv:2401.15365}, 2024.

\bibitem{tie2025survey}
Guiyao Tie, Zeli Zhao, et~al.,
\newblock ``A survey on post-training of large language models,''
\newblock {\em arXiv e-prints}, pp. arXiv--2503, 2025.

\bibitem{deepseek-r1}
Daya Guo, Dejian Yang, et~al.,
\newblock ``Deepseek-r1: Incentivizing reasoning capability in llms via reinforcement learning,''
\newblock {\em arXiv preprint arXiv:2501.12948}, 2025.

\bibitem{Wang2025InternVL35}
Weiyun Wang, Zhangwei Gao, et~al.,
\newblock ``Internvl3.5: Advancing open-source multimodal models in versatility, reasoning, and efficiency,'' 2025.

\bibitem{zhang2025acl}
Ruohong Zhang, Bowen Zhang, et~al.,
\newblock ``Improve vision language model chain-of-thought reasoning,''
\newblock {\em arXiv preprint arXiv:2410.16198}, 2024.

\bibitem{qwen2_5_vl}
Shuai Bai, Keqin Chen, et~al.,
\newblock ``Qwen2. 5-vl technical report,''
\newblock {\em arXiv preprint arXiv:2502.13923}, 2025.

\bibitem{EVOBC}
Haisu Guan, Jinpeng Wan, et~al.,
\newblock ``An open dataset for the evolution of oracle bone characters: Evobc,''
\newblock {\em arXiv preprint arXiv:2401.12467}, 2024.

\bibitem{OracleSage}
Hanqi Jiang, Yi~Pan, et~al.,
\newblock ``Oraclesage: Towards unified visual-linguistic understanding of oracle bone scripts through cross-modal knowledge fusion,''
\newblock {\em arXiv preprint arXiv:2411.17837}, 2024.

\bibitem{CFIRN}
Zhicong Wu, Qifeng Su, et~al.,
\newblock ``A cross-font image retrieval network for recognizing undeciphered oracle bone inscriptions,''
\newblock {\em arXiv preprint arXiv:2409.06381}, 2024.

\bibitem{wang2025simplear}
Junke Wang, Zhi Tian, et~al.,
\newblock ``Simplear: Pushing the frontier of autoregressive visual generation through pretraining, sft, and rl,''
\newblock {\em arXiv preprint arXiv:2504.11455}, 2025.

\bibitem{li2024look}
Yian Li, Wentao Tian, et~al.,
\newblock ``Look before you decide: Prompting active deduction of mllms for assumptive reasoning,''
\newblock {\em arXiv preprint arXiv:2404.12966}, 2024.

\bibitem{liu2025noisyrollout}
Xiangyan Liu, Jinjie Ni, et~al.,
\newblock ``Noisyrollout: Reinforcing visual reasoning with data augmentation,''
\newblock {\em arXiv preprint arXiv:2504.13055}, 2025.

\bibitem{chen2025compile}
Zuyao Chen, Jinlin Wu, et~al.,
\newblock ``Compile scene graphs with reinforcement learning,''
\newblock {\em arXiv preprint arXiv:2504.13617}, 2025.

\bibitem{wang2025skywork}
Peiyu Wang, Yichen Wei, et~al.,
\newblock ``Skywork r1v2: Multimodal hybrid reinforcement learning for reasoning,''
\newblock {\em arXiv preprint arXiv:2504.16656}, 2025.

\bibitem{zhang2025adamhf}
Shuaiyu Zhang, Xun Lin, et~al.,
\newblock ``Adamhf: Adaptive multimodal hierarchical fusion for survival prediction,''
\newblock {\em arXiv preprint arXiv:2503.21124}, 2025.

\bibitem{Li2025ThinkOrNotThink}
Ming Li, Jike Zhong, et~al.,
\newblock ``Think or not think: A study of explicit thinking in rule-based visual reinforcement fine-tuning,'' 2025.

\bibitem{KimiTeam2025KimiVL}
Kimi Team, Angang Du, et~al.,
\newblock ``Kimi-vl technical report,'' 2025.

\bibitem{KimiTeam2025KimiK15}
Kimi Team, Angang Du, et~al.,
\newblock ``Kimi k1.5: Scaling reinforcement learning with llms,'' 2025.

\bibitem{Shen2025VLMR1}
Haozhan Shen, Peng Liu, et~al.,
\newblock ``Vlm-r1: A stable and generalizable r1-style large vision-language model,'' 2025.

\bibitem{Yu2025PerceptionR1}
En~Yu, Kangheng Lin, et~al.,
\newblock ``Perception-r1: Pioneering perception policy with reinforcement learning,'' 2025.

\bibitem{Chen2025VisRL}
Zhangquan Chen, Xufang Luo, et~al.,
\newblock ``Visrl: Intention-driven visual perception via reinforced reasoning,'' 2025.

\bibitem{Deng2025OpenVLThinker}
Yihe Deng, Hritik Bansal, et~al.,
\newblock ``Openvlthinker: Complex vision-language reasoning via iterative sft-rl cycles,'' 2025.

\bibitem{v-oracle}
Runqi Qiao, Qiuna Tan, et~al.,
\newblock ``V-oracle: Making progressive reasoning in deciphering oracle bones for you and me,''
\newblock in {\em Proceedings of the 63rd Annual Meeting of the Association for Computational Linguistics (Volume 1: Long Papers)}, 2025, pp. 20124--20150.

\bibitem{qwen3}
An~Yang, Anfeng Li, et~al.,
\newblock ``Qwen3 technical report,''
\newblock {\em arXiv preprint arXiv:2505.09388}, 2025.

\bibitem{focalpo}
Tong Liu, Xiao Yu, et~al.,
\newblock ``Focalpo: Enhancing preference optimizing by focusing on correct preference rankings,''
\newblock {\em arXiv preprint arXiv:2501.06645}, 2025.

\end{thebibliography}

\end{document}